\documentclass[DIV=calc, paper=letter, fontsize=11pt]{scrartcl}	 

\usepackage[]{units}

\usepackage{lipsum} 
\usepackage[english]{babel} 
\usepackage[protrusion=true,expansion=true]{microtype} 
\usepackage{amsmath,amssymb,amsfonts,amsthm,bm,bbm} 

\usepackage[svgnames, table]{xcolor} 

\usepackage[hang, small,labelfont=bf,up,textfont=it,up]{caption} 
\usepackage{booktabs} 

\usepackage{fix-cm}
\usepackage[]{units}

\usepackage{subfig}
\usepackage{physics}

\usepackage{tabularx}

\usepackage[colorlinks=true, linkcolor=blue, citecolor=blue, filecolor=blue,
runcolor=blue, urlcolor = blue]{hyperref} 

\usepackage[capitalize, noabbrev]{cleveref}

\usepackage{fullpage}
\usepackage{sectsty} 

\usepackage{fancyhdr} 
\usepackage{lastpage}

\usepackage{nicefrac}

\usepackage{tikz}
\usetikzlibrary{calc,patterns,decorations.pathmorphing,decorations.markings}
\usepackage{pgfplots}

\usepackage{multicol}

\usepackage{float}

\usepackage[utf8]{inputenc}
\usepackage[english]{babel}
\usepackage{tikz}
\usepackage{comment}

\usetikzlibrary{arrows}

\usepackage{mathtools} 

\usepackage{algorithm, algpseudocode}

\DeclareMathOperator*{\argmin}{arg\,min}

\usepackage[
backend=bibtex, 
style=numeric-comp,
sorting=none
]{biblatex}

\makeatletter
\newcommand{\multicolinterrupt}[1]{
\setcounter{tempcolnum}{\col@number}
\end{multicols}
#1
\begin{multicols}{\value{tempcolnum}}
}
\makeatother

\definecolor{issuePJA_color}{rgb}{1.0,0.0,0.0}

\definecolor{commentPJA_color}{rgb}{1.0,0.0,0.8}

\newcommand{\commentPJA}[1]{}

\definecolor{issueQB_color}{rgb}{1.0,0.8,0.0}

\newcommand{\issueQB}[1]{}

\definecolor{commentQB_color}{rgb}{0.6,0.0,0.8}

\definecolor{rev_color}{rgb}{0.0,0.0,0.0}

\newcommand{\rev}[1]{{\textcolor{rev_color}{#1}}}

\definecolor{atz_table1}{rgb}{0.85, 0.85, 0.85}
\definecolor{atz_table2}{rgb}{0.8, 0.8, 0.8}
\definecolor{atz_table3}{rgb}{0.75, 0.75, 0.75}

\newcommand{\mb}[1]{\mathbf{#1}}
\newcommand{\bsy}[1]{\boldsymbol{#1}}

\newcommand{\percent}{\%}

\newcommand{\FirstForm}{\bsy{\mathrm{I}}}
\newcommand{\SecondForm}{\bsy{\mathrm{I\!I}}}

\newcommand{\R}{\mathbb{R}}

\newcommand{\pn}[1]{\left( #1 \right)}
\newcommand{\bk}[1]{\left[ #1 \right]}
\newcommand{\set}[1]{\left\{ #1 \right\}}

\usepackage{graphicx}
\DeclareGraphicsExtensions{.png}

\usepackage{lettrine} 
\newcommand{\initial}[1]{ 
\lettrine[lines=3,lhang=0.3,nindent=0em]{
\color{DarkGoldenrod}
{\textsf{#1}}}{}}

\usepackage{caption}

\DeclareOldFontCommand{\bf}{\normalfont\bfseries}{\mathbf}

\usepackage{graphicx}

\usepackage{titling} 

\newcommand{\HorRule}{\color{DarkGoldenrod}

\rule{\linewidth}{1pt}} 

\pretitle{\vspace{-80pt} \begin{flushleft} \HorRule \fontsize{12}{12}
\color{DarkRed} \selectfont} 

\title{Transferable Foundation Models for Geometric Tasks on Point Cloud Representations: Geometric Neural Operators}

\posttitle{\par\end{flushleft}}

\preauthor{\vspace{-15pt} \begin{flushleft}\fontsize{12}{12} 
\color{DarkRed}} 
\author{B. Quackenbush$^{1}$, P. J. Atzberger$^{1,2}$ }

\postauthor{\small \color{Black} 
\\ $[1]$ Department of Mathematics,
University of California Santa Barbara (UCSB). \\
$[2]$ (corresponding author), Department of Mathematics, Department of 
Mechanical Engineering, 
University of California Santa Barbara (UCSB); atzberg@gmail.com;
\url{http://atzberger.org/}
\vskip 1.5em

\vspace{-0.6cm}
\HorRule
\end{flushleft} 
} 

\date{\today}

\pgfplotsset{compat = 1.16}

\DeclareGraphicsExtensions{.png}

\begin{document}
\date{}
\maketitle

\thispagestyle{fancy}

\vspace{-1.75cm}

\initial{W}\textbf{\rev{e introduce methods for obtaining pretrained Geometric
Neural Operators (GNPs) that can serve as basal foundation models for use in
obtaining geometric features.  These can be used within data processing
pipelines for machine learning tasks and numerical methods.  We show how our
GNPs can be trained to learn robust latent representations for the differential
geometry of point-clouds to provide estimates of metric, curvature, and other
shape-related features.  We demonstrate how our pre-trained GNPs can be used
(i) to estimate the geometric properties of surfaces of arbitrary shape and
topologies with robustness in the presence of noise, (ii) to approximate
solutions of geometric partial differential equations (PDEs) on manifolds, and
(iii) to solve equations for shape deformations such as curvature driven flows.
We release codes and weights for using GNPs in the package
\href{https://github.com/atzberg/geo_neural_op}{geo\_neural\_op}.  This allows for incorporating
our pre-trained GNPs as components for reuse within existing and new data
processing pipelines.  The GNPs also can be used as part of numerical solvers
involving geometry or as part of methods for performing inference and other
geometric tasks.  } }

\setlength{\parindent}{5ex}

\section*{Introduction} 
A recent development in machine learning has been increasing efforts to
formulate and train models for reuse across a broad range of related tasks.
These are often referred to as \textit{foundation models} to indicate they are
to be built upon in order to perform further
tasks~\cite{bommasani2021opportunities,standfordblog2021}.  Prominent recent
examples include large language models
(LLMs)~\cite{touvron2023llama,koroteev2021bert,ChatGPT2023},
image  models~\cite{DallE2024,podell2023sdxl,Midjourney2024},
and object detection
models~\cite{redmon2016you,carion2020end,minderer2022simple}.
Such approaches highlight the utility of having off-line training protocols and
models with capabilities that are transferable to facilitate a wide range of
tasks.  Further specialized developments of shared off-line models even at
smaller computational scales hold promise for impacting fields that include
inference in scientific machine learning, physics-based feature extraction, and
developing numerical solvers and
simulations~\cite{atzberger2018importance,baker2019workshop,
Quackenbush2024,Atzberger2024sdyngans}.

For geometric tasks, we introduce methods for developing pre-trained basal
foundation models that are transferable for tasks involving point-clouds and
other types of representations.  Our approaches allow for
discovering geometric structures and processing methods without the need for
meshing or retraining to obtain information for down-stream tasks.  We build on
our work on Geometric Neural Operators (GNPs) in \cite{Quackenbush2024}.  Our
methods can be used to obtain estimates that include different types of
curvatures, metrics, and other geometric features.  Our methods also can be
used as part of evaluation of differential operators on manifold surfaces,
exterior calculus operations, and other procedures that arise in geometric
analysis and numerical solvers for geometric
PDEs~\cite{Atzberger2018,gonzalez2008survey,rower2022surface,millan2013nonlinear}.
Our training methods also incorporate approaches for obtaining estimates that
are robust to noise or designed to deal with artifacts that can arise in
point-cloud representations and other datasets.  Our methods allow for learning
pre-trained GNP models that can then be deployed within other computational
methods and data-processing pipelines to perform geometric tasks. 

Geometric problems arise in many inference settings and numerical methods
\cite{Bronstein2017,Wasserman2018,Meilua2023,Ortiz2016}.  Related 
challenges include approaches for handling point-cloud representations in shape
classification, segmentation, and 
~\cite{Hackel2017,Atzberger2022,zaheer2017deep,
qi2017pointnet,shapenet2015,mullen2010signing,guo2020deep},
developing PDE solvers on manifolds~\cite{Atzberger2022,Atzberger2020,serrano2023operator,
Atzberger2018,budd1999geometric,arroyo2007local,liang2013solving,dziuk2013finite}, 
and geometric processing of meshes, lidar measurements, and other types of data
\cite{wang2015designing,wang2015designing,li2020deep,alaba2022survey,
wang2019dynamic,ruoppa2025unsupervised,mullen2010signing}.  Related prior work includes 
development of neural operators for processing point-clouds  
in~\cite{pang2024neural,pang2023learning}.  In this work, graph neural networks are 
trained to mimic the action of the Laplace-Beltrami operator for use
in estimation of local geometric properties, geodesic distances, 
deformations, and spectral reductions~\cite{pang2024neural,pang2023learning}.  
Related work on approximating solutions of 
Partial Differential Equations (PDEs) on manifolds
based on machine learning methods and meshfree methods
include~\cite{zeng2025point,Atzberger2020,serrano2023operator,
arroyo2007local,Atzberger2022,liang2024solving,li2023geometry,
chen2024learning,zhong2025physics,zhao2025diffeomorphism,franco2023mesh}. 
In these methods, estimators are developed for meshes or
point-cloud representations of the manifolds to obtain 
geometric quantities used in evaluation of the 
differential operators.  In~\cite{zeng2025point,li2023fourier,
serrano2023operator,li2023geometry,chen2024learning,yin2024scalable},
more general neural operators are trained to learn surrogate models
for parametric PDEs as part of solving inverse 
and design problems.  The methods are trained primarily for 
processing data globally to obtain operators that 
learn implicit latent representations 
for making predictions for solutions over a parameterized class of 
shapes or PDEs.  These methods have been shown to be useful for geometries
and functions close to the training datasets, and to help in improving the efficiency of
methods used in inverse and design problems.  However, since the learned
representations are implicit, this poses some limits to transferability 
to other geometries and problem settings.  Another important issue that arises in
practice for geometries obtained from data-driven methods or 
empirical measurements is the need for robustness to sampling noise and 
other artifacts.

We address these and other challenges by developing methods building on 
Geometric Neural Operators (GNPs)~\cite{Quackenbush2024}.  This 
provides the basis for obtaining robust pre-trained
transferable foundation models for diverse geometric tasks.  We demonstrate use
cases for our GNPs showing their robustness, transferability, and 
characterize their other properties.
This includes (i) performing validation studies against test sets involving
topologies and geometric shapes not used in training, (ii) computing
deformations of shapes driven by mean-curvature flows, and (iii) developing
geometric PDE solvers based on the GNP models.  The results indicate the
pre-trained GNPs can be used as data-driven alternatives to development of 
hand-crafted geometric estimators, which are often technical to formulate
analytically and implement in practice.  Our data-driven approaches also 
provide ways to enhance the robustness of estimators by allowing for 
noise and other artifacts to be taken into account during training.  
We release our codes and model
weights in a package \url{https://github.com/atzberg/geo_neural_op}.
Our package can be used both for training new models or for 
incorporating our pre-trained GNP models into other data 
processing pipelines and computational methods. 
Our training approaches provide
ways to learn data-driven filtering and adjustments for 
handling noise and other artifacts in point-clouds.  The
methods can be used to obtain transferable 
pre-trained models that can be used for 
performing diverse geometric tasks.

We organize our paper as follows.  We discuss how geometric features can be
learned from point cloud representations using Geometric Neural Operators
(GNPs) in Section~\ref{sec_learning_features}.  We discuss our approach to obtain
transferable pre-trained GNPs for geometric tasks in
Section~\ref{sec_training_gnps}.  We discuss results using our transferable
GNP models in Section~\ref{sec_results}.  This includes demonstrating 
the methods for 
(i) estimating metrics and curvatures of surfaces, 
(ii) shape changes driven by mean-curvature flows, and 
(iii) development of numerical methods for solving geometric PDEs.

\section{Geometric Neural Operators (GNPs) for Point-Cloud Representations}
\label{sec_learning_features}

Geometry plays an important role in many machine learning and
numerical tasks.  This includes classification, segmentation, and  of
shapes \cite{Hackel2017,Atzberger2022,zaheer2017deep,
qi2017pointnet,shapenet2015,mullen2010signing,guo2020deep}, 
approximation of solutions of PDEs on manifolds
\cite{Atzberger2020,serrano2023operator,
Atzberger2018,arroyo2007local,dziuk2013finite}, and
other tasks
\cite{wang2015designing,wang2015designing,li2020deep,alaba2022survey,
wang2019dynamic,ruoppa2025unsupervised}.  For point-cloud data,
a fundamental challenge is to estimate reliably the 
curvature and other intrinsic geometric quantities from discrete 
samples of the shape.  This can be especially challenging in the 
presence of noise and other artifacts that can arise in practice.

We develop neural operators for learning representations from the point-cloud
data building on~\cite{Kovachki2023a,Chen1995} and our prior work introducing
geometric neural operators (GNPs) in~\cite{Quackenbush2024}.  In contrast to
conventional neural networks, neural operators provide a distinct class of
models that can learn mappings between function spaces that are independent of
the specific underlying discretizations.  The key idea is to model sample
evaluations of the input function as a collection of features and to learn
various types of integral and linear operations that successively process the
features of the input.  The methods allow for readily recovering Fourier
transform-based methods, kernel integral operator methods, and other non-linear
transformations of functions~\cite{Kovachki2023a,Chen1995,Quackenbush2024}.
This provides a powerful framework for learning operations on diverse classes
of functions.  We utilize this approach to learn mappings from point-cloud
samplings of the geometry to functions capturing geometric
features~\cite{Quackenbush2024}.

\begin{figure}
\centering
\includegraphics[width=0.99\linewidth]{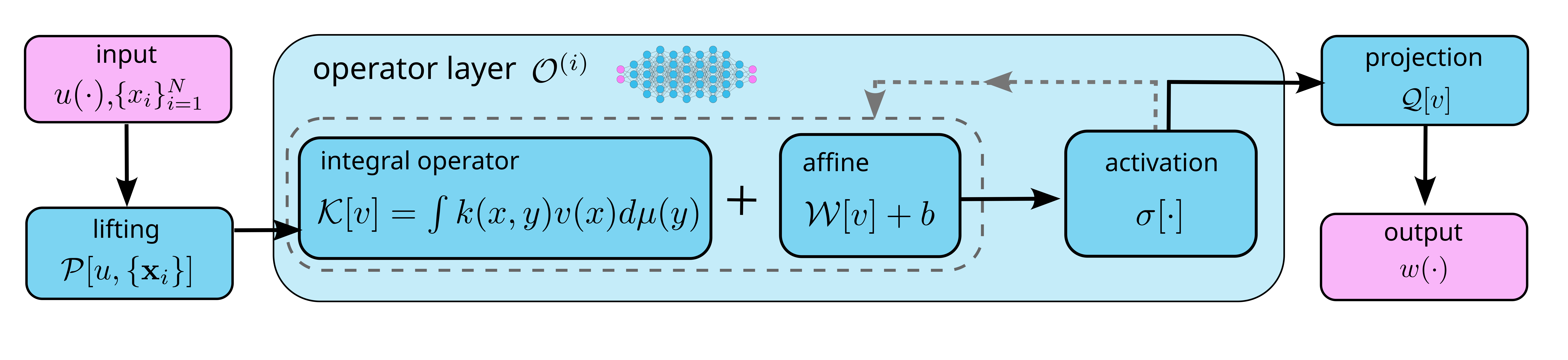}
\caption{\textbf{Geometric Neural Operators (GNPs).} 
We learn features from point-clouds  using geometric neural operators
GNPs~\cite{Quackenbush2024}.  These neural operators consist of three learnable
components (i) a lifting operation $\mathcal{P}[u,\{\mb{x}_i\}]$ that provides initial
features for the input
geometry data $\{\mb{x}_i\}_{i=1}^N$ and 
input function $u(\cdot)$, (ii) layers of
operators $\mathcal{O}^{(i)}$ that consist of 
kernel integral operations $\mathcal{K}[v](x)$ and 
affine operations $\left(\mathcal{W}v\right)(x) + b(x)$ each of which
are passed through a non-linear activation operation $\sigma[\cdot]$, 
and (iii) a projection operation $\mathcal{Q}$ for constructing the
final output function $w(x) = w(\xi^1,\xi^2)$ and local 
parameterization $\bsy{\tilde{\sigma}}(\xi^1,\xi^2) = \mb{\bar{x}} + \xi^1 \psi_1 + \xi^2\psi_2$.
We use geometric neural operators (GNPs) to map a
collection of points $\{x_i\}_{i=1}^N$ sampled from the geometry to local 
parameterizations and functions $w = w(\xi^1,\xi^2)$.  
}
\label{fig_gnp}
\end{figure}

\subsection{Learning Latent Geometric Features with GNPs}
\label{sec_learning_gnps}

We develop methods building on Geometric Neural Operators (GNPs) introduced 
in \cite{Quackenbush2024}.
GNPs represent operators  
$\mathcal{G}_\theta: \mathcal{A} \to \mathcal{U}$, where $\mathfrak{a} \in \mathcal{A}$,
$\mathfrak{a} = (u(\cdot),\Phi(\cdot))$ with $u(\cdot)$ is a function and  
$\Phi(\cdot)$ denotes geometric information.  Many choices can be used for 
$\Phi$ to provide a description of the
manifold shape $\mathcal{M}$.  Here, we consider the collection of points within 
an embedding in $\mathbb{R}^n$ of $\mathcal{M}$. 
In practice, we use for $\mathfrak{a}$ a discretization $a$ where
$a \in \R^{d_a}$ and $a = \{u(\tilde{x}_i)\}_{i=1}^N,\{x_i\}_{i=1}^N$,
where $u(\tilde{x}_i)$ denote function evaluations and 
$x_i$ denote sample points in $\mathcal{M}$.  Here we allow for the case 
where $\tilde{x}_i \neq x_i$. 

We use GNPs that consist of the following three
learnable components, (i) a lifting operator $\mathcal{P}$ for $a
\in \R^{d_a}$ where $a = \{u(\tilde{x}_i)\}_{i=1}^N,\{x_i\}_{i=1}^N$ samples
function evaluations
to a higher dimensional set of feature functions $v_0 \in
\R^{d_v}$ with $d_v \geq d_a$, (ii) a composition of layers
consisting of integral operators $\mathcal{K}[v]$ and 
affine operators $\mathcal{W}[v](x) + b(x)$ that are processed 
by a non-linear activation $\sigma[\cdot]$ to obtain
$v_{i+1} = \sigma \pn{Wv_i + \mathcal{K}[v_i] + b}$, and (iii) a
projection operator $\mathcal{Q}$ to construct an $\R^{d_u}$-valued function
$w$.  The trainable sub-components also include within 
the operator layers the kernel $k$, bias function $b$, and local 
function operator $W$.  We show the parts of the geometric 
neural operator in Figure~\ref{fig_gnp}.
 This gives a neural operator with $T$ layers
of the general form
\begin{equation} 
    \mathcal{G}_\theta^{(T)} = \mathcal{Q}
\circ \sigma_T\pn{W_T + \mathcal{K}_T + b_T} \circ \dots \circ \sigma_0\pn{W_0
+ \mathcal{K}_0 + b_0} \circ \mathcal{P}.  
\end{equation}
We collect all trainable
parameters into $\theta$.
The activation of the last layer $\sigma_T$ is typically taken to be the identity.

We consider linear operators $\mathcal{K}$ of the form
\begin{equation}\label{kernel_integral}
\mathcal{K}[v](x) = \int_D k(x, y) v(y) \ d\mu(y).
\end{equation}
The $\mu$ is a measure on $D \subset \R^{d_{v}}$, $v:D \to \R^{d_{v}}$ is the
input function, and $k$ is a kernel $k(x, y) \in \R^{d_{v}} \times \R^{d_{v}}$.
In practice, we approximate this integral on a truncated domain $B_r(\bsy x)$ using 
\begin{equation}
    \label{equ_kernel_approx}
    \tilde{\mathcal{K}}[v_t](\bsy x_j) = 
    \frac{1}{N} \sum_{x_k \in B_r(\bsy x_j)} k(x_j, x_k) v_t(x_k).
\end{equation}
This discretization can be interpreted as message passing on a
graph~\cite{gilmer2020message,gori2005new}. 
For each layer $t$, we use a trainable kernel $k = k(x,y;\theta_t)$
parameterized by $\theta_t$ for fully connected neural networks having
layer-widths $(d_a, n, d_v^2)$ for $n \in \mathbb{N}$. We increase efficiency by
using Nystrom approximation, summing over a subset of the points in $B_r(\bsy x_j)$
instead of all points, using a maximum of $32$ points in the sum. We also enforce
a block-factorized structure in the outputs of the kernel network $k$, as in~\cite{Quackenbush2024}. 
To obtain the output function representation, we apply 
average pooling to the outputs of $\mathcal{G}_\theta$ and pass
this through two fully connected layers to obtain our final output 
functions. For additional discussions of GNPs and technical 
details, see \cite{Quackenbush2024}.

We use GNPs to learn latent representations of the geometry from point cloud
representations $\mathcal{X} = \{\mb{x}_i\}_{i=1}^M$ of a manifold
$\mathcal{M}$.  In general, we consider samplings where $\mb{x}_i =
\mb{\hat{x}}_i + \zeta_i$ with $\mb{\hat{x}}_i \in \mathcal{M}$ and $\zeta_i$
is noise obscuring the geometry.  For each point $\mb{x} \in \mathcal{X}$, we
define the neighborhood $\mathcal{N_{\epsilon}}(\bsy x) = \{\mb{x}_i \;| \;
\|\mb{x}_i - \mb{x}\| < \epsilon \}$ as the set of points within distance
$\epsilon$ of $\mb{x}$.  To obtain good characteristic scales, we take
$\epsilon = r_k\tau$, $r_k$ to be the radius of the $k$th nearest neighbor of
$\bsy x$, and $\tau$ as a user-specified parameter. Typical values we use are
$\tau = 1.1$ and $k \in \set{30, 50}$.  

As part of our GNPs for point-cloud processing, we output for 
a neighborhood $\mathcal{N}_\epsilon$ around each point $\mb{x}$
a local coordinate system $(\xi^1,\xi^2)$.  We construct this
by performing the following computations as part of our GNPs.
We center the points using $\mathcal{N}_\epsilon - \bar{\mb{x}}$
where $\bar{\mb{x}}$ is the mean and we perform Principal Component Analysis
(PCA)~\cite{hastie2009elements,hotelling1933analysis,pearson1901liii}.  
This provides a local orthonormal basis $\{\bsy{\psi}_j\}_{j=0}^3$ such that
each $\bsy y \in \mathcal{N}_\epsilon(\bsy x)$ can be expressed as
\begin{equation}
\label{equ_local_pca}    
\bsy y = \bar{\bsy x} 
+ \xi^1(\bsy y) \bsy \psi_1 
+ \xi^2(\bsy y) \bsy \psi_2 
+ \xi^3(\bsy y) \bsy \psi_3.
\end{equation}
The $\xi^j(\mb{y})$ provide a representation of $\mb{y}$ in terms of this
basis.  In the limit of increasingly dense samplings of $\mathcal{M}$, the
vectors $\bsy{\psi}_1$, $\bsy{\psi}_2$ would approach tangent vectors of the
manifold and the vector $\bsy{\psi}_3$ approaches the normal vector.  For
finite samplings these provide approximations for obtaining a local coordinate
system.  We set these vectors to have unit norm, $\|\bsy{\psi}_j\| = 1$.  To
ensure the correct orientation of the coordinate frame, we assume the manifold
has a known orientation on $\mathcal{N_\epsilon}(\bsy x)$, and we setup the
basis so $\bsy \psi_1 \times \bsy \psi_2 = \bsy \psi_3$ is aligned with the
outward normal at $\bar{\bsy x}$.  We remark these procedures and PCA 
also can be replaced by learning alternative operators during 
training to obtain other representations and coordinate systems. 

We further canonicalize the point data for our training methods by using
transformations that impose invariances to rotations and translations, and
equivariances to scalings.  The invariance to rotations and translations is
accomplished by using the representation $\xi^j(\mb{y})$ obtained from the
basis $\bsy{\psi}_j$.  In particular, we use the change of coordinates from
$\mb{y} = (y_1,y_2,y_3)$ to to $\bsy{\xi} = (\xi^1,\xi^2,\xi^3)$ over basis
$\set{\bsy \psi_j}$.  We also perform further scalings of $\xi^j$ to normalize
the point data, which we discuss below.  Our normalizations allow our learning
methods to find common patterns and features spanning a broad range of similar
geometries.  Our approach provide ways to amplify the statistical power of the
point-cloud training datasets for learning operations.

An important inductive bias we incorporate into our methods is to find a local
Monge-Gauge parametrization of the surface patch $\mathcal{N}_\epsilon(\bsy x)$
\cite{monge1809application,pressley2010elementary}.  We use parameterizations
of the form 
\begin{equation}
\label{equ_monge-gauge}    
\bsy \sigma(\xi^1, \xi^2) = 
\bar{\bsy x} 
+ \xi^1 \bsy \psi_1 
+ \xi^2 \bsy \psi_2 
+ h(\xi^1, \xi^2) \bsy \psi_3.
\end{equation}
This requires finding a height function $h(\xi^1, \xi^2)$ that matches
$\xi^3(\mb{y})$ for each sample point $\mb{y} \in \mathcal{N}_\epsilon$.  We
can then use $\bsy{\sigma}$ as part of extracting local geometric quantities,
contributions to the action of differential operators, and other features.  
We show an example in Figure~\ref{fig_gaussian_curvature}.  We
give explicit expressions for common geometric quantities and operators 
that can be obtained from $\bsy{\sigma}$ and $h$ in Appendix~\ref{appendix_A}.  

\begin{figure}
\centering
\includegraphics[width=0.99\linewidth]{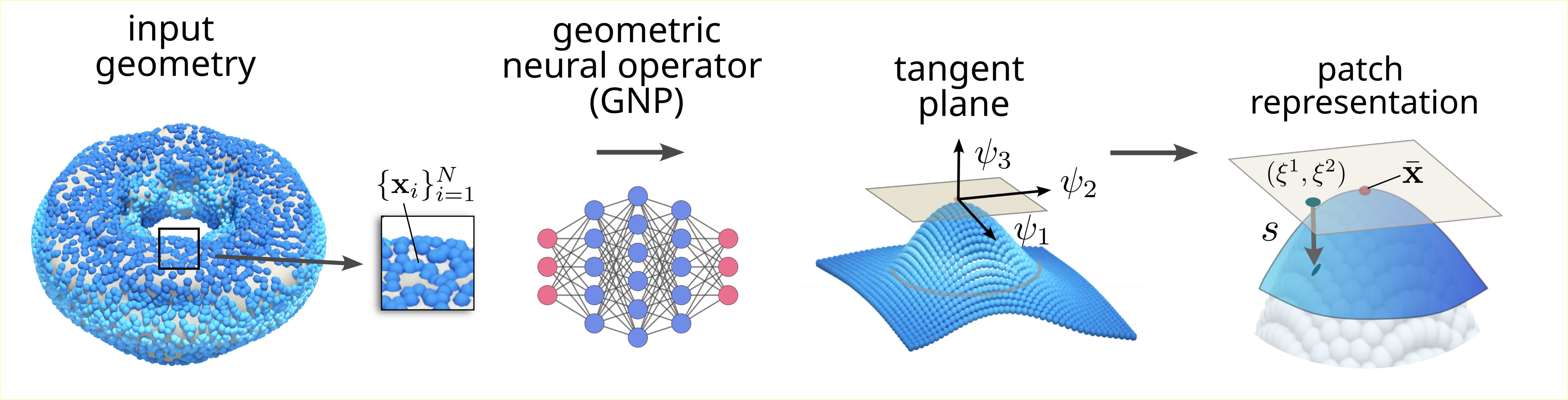}
\caption{\textbf{Learning Latent Geometric Representations.} 
We show how data is processed by our geometric neural operators (GNPs).
We obtain latent representations at each $\mb{\bar{x}}$ by learning 
GNPs that map a collection of points $\{\mb{x}_i\}_{i=1}^N$ sampling the geometry 
to a local parameterization $(\xi^1,\xi^2)$ and surface height function $s(\xi^1,\xi^2)$. 
This provides geometric information for further processing and tasks.
}
\label{fig_rep_learning}
\end{figure}

To further normalize the data, we perform additional scalings to obtain results
more robust to manifolds having different local characteristic scales.  While
scales are associated with important geometric properties, we utilize that
geometric quantities satisfy many types of equivariances allowing for further
canonicalization in terms of more intrinsic features of the geometry.  We use
the following canonical rescaling of the coordinates
$(\tilde{u},\tilde{v},\tilde{w}) = (\xi^1,\xi^2,\xi^3)$ to $(u,v,w) = (
\epsilon\tilde{u}, \epsilon\tilde{v}, \delta\tilde{w})$. The $\epsilon$ and $\delta$ 
give characteristic scales that serve to standardize the representation of the
local geometry.   We take  $\epsilon$ as the radius of the neighborhood
$\mathcal{N}_\epsilon$.  We obtain $\delta$ by $\delta = \max\pn{2 \lambda,
\delta_0}$, where $\lambda$ is the standard deviation in the $\bsy \psi_3$
direction, and $\delta_0$ is a user-specified parameter.  This yields a
rescaled Monge-Gauge parameterization of the form
\begin{equation}
\label{equ_patch_eq}
\bsy \sigma(u, v) = 
\bar{\bsy x} 
+ u \bsy \psi_1 
+ v \bsy \psi_2 
+ s(u, v) \bsy \psi_3.
\end{equation}
This gives the rescaled height $s(u,v) = \delta\cdot h(\xi^1,\xi^2)$.
Our rescaling approach provides in the geometric setting 
methods similar to 
batch-normalization~\cite{goodfellow2016deep,bjorck2018understanding} 
to provide a more uniform set of scales to enhance learning.  

An important issue that arises in practice is that we need to be cautious in  
extreme cases when $\mathcal{N}_\epsilon(\bsy x)$ is nearly flat.  In
this case, $\lambda$ is nearly zero so $s(u, v)$ is prone to fit
noise in the data or numerical round-off errors.  To mitigate this issue, 
we introduce the user-specified parameter $\delta_0$ and set it 
in practice to be $0.005$ as a default value.  

To further filter noise during processing, we limit the capacity of the 
height function in equation~\ref{equ_patch_eq}.  We use in the final
projection operations the functional form
\begin{eqnarray}
\label{equ_legendre}
s(u,v) = \sum_k s_k \phi_k(u,v). 
\end{eqnarray}
The choice of $\Phi = \set{\phi_1, \dots, \phi_N}$ allows for
different ways to locally filter the geometry.
We use a set of  
basis functions that are tensor products of the form
$    \Phi = \set{l_i(u)l_j(v)}_{i, j=0}^{d},$
where $l_i$ give the basis set of functions. We use for this purpose 
the orthogonal Legendre Polynomials~\cite{boyd2001chebyshev,Trefethen2000}. 
By taking Taylor expansions for smooth surfaces,  we see any shape can be well
approximated locally with this choice by taking sufficiently small
neighborhoods $\mathcal{N}_\epsilon(\bsy x)$. In practice, we use $d=3$ for
approximating $s(u,v)$ which serves to help further filter out noise and
over-fitting.  We use this as part of the GNP mappings 
from the input point-cloud data
$\{\mb{x}_i\}_{i=1}^N$ to obtain at 
each $\mb{\bar{x}}$ the local latent representations $s(u,v)$ for the geometry.  
This allows for the GNP training methods to obtain latent geometric representations
that can handle data containing noise and other 
artifacts.  We show the latent representations  
in Figure~\ref{fig_rep_learning}.

\section{Training Transferable GNP Models for Geometric Tasks}
\label{sec_training_gnps}

\begin{figure} 
\centering
\includegraphics[width=0.99\linewidth]{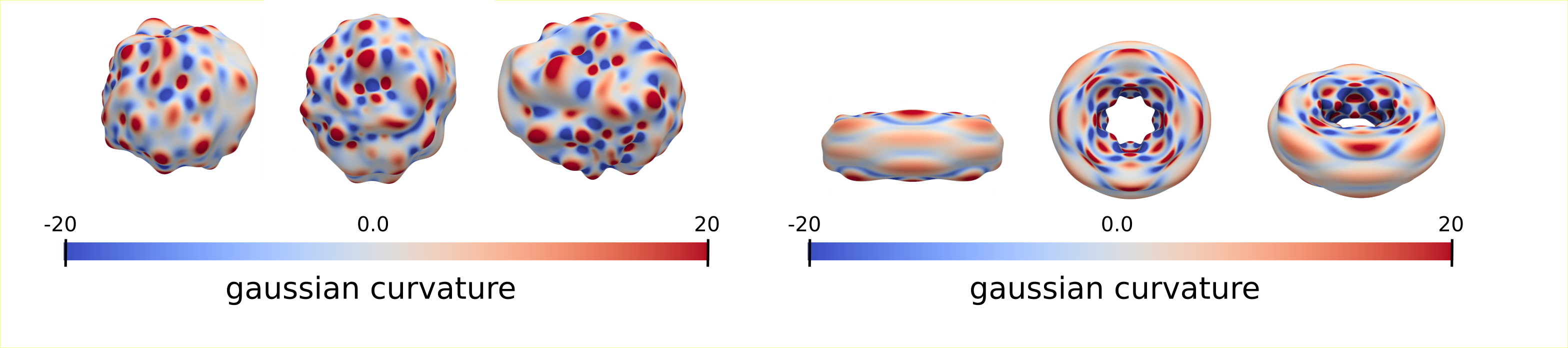}
\caption{\textbf{GNP Estimators for Gaussian Curvatures on Test Shapes.}
Gaussian curvatures of radial manifold  \textit{(left)} and toroidal manifold
\textit{(right)}.  The views show the same manifold from a few different vantage points.} 
\label{fig_gaussian_curvature} 
\end{figure}

We now show how GNPs can be trained to obtain transferable models for geometric
tasks.  We start by training GNPs by leveraging geometric information readily
available from the parameterizations of radial manifolds computed
using spectral numerical methods and
spherical harmonics~\cite{Atzberger2018}. We show an example 
radial manifold in Figure~\ref{fig_gaussian_curvature}.
We then show how the trained GNPs can then
be used for other more general geometries and topologies.  

For training, we consider radial manifolds which consist of shapes having the
star property that each point on the surface can be connected by a line segment
to the origin without self-intersections.  Radial manifolds can be
parameterized by two coordinate charts in spherical coordinates in the 
form~\cite{Atzberger2018,sigurdsson2016hydrodynamic} 
\begin{equation} \label{equ_radial_manifold}
\bsy{x}(\theta, \phi) = \bsy \sigma (\theta, \phi) = r(\theta, \phi) \bsy 
r(\theta, \phi).
\end{equation}
The $\bsy r(\theta,\phi)$ is the unit vector from the origin to the point on
the sphere with spherical coordinates $(\theta, \phi)$.  The $r(\theta,\phi)$
is a positive scalar function giving the distance from the origin. 
For analysis and numerical computations, we use spherical harmonics expansions
of   
\begin{equation}
\label{equ_radial_function}
r(\theta, \phi) = \sum_{l=0}^{L} \sum_{m=-l}^{l} a_l^m Y_l^m(\theta, \phi),
\end{equation}
where $Y_l^m$ are the spherical harmonic of index $m$ and order $l$. 
We compute expressions numerically using the package sphericart 
\cite{sphericart} for the spherical harmonics and 
our approaches in~\cite{Atzberger2018,sigurdsson2016hydrodynamic}. 

We generate shapes for our training datasets of varying complexity by adjusting
$L$ in our spherical harmonics expansions for the radial functions.  We obtain
for $r(\theta,\phi)$ the complex coefficients $a_l^m = \alpha_l^m + i\beta_l^m$
for $m = 0, \dots, l$, where the real $\alpha_l^m$ and imaginary $\beta_l^m$,
by sampling the components from normal distributions with mean $0$ and standard
deviation $\tfrac{1}{l}$.  To ensure real-valued functions we generate
coefficients for $a_l^m$ for $m< 0$ and set the other coefficients using
conjugacy conditions.  To vary the shapes of the geometry, we also select
different values of $L$ ranging over $L \in \{3, 6, 8, 10, 12, 15, 18, 22 \}$.
As another convention in generating our training datasets, we also  translate
and scale $r(\theta, \phi)$ so that it has mean $1$ and satisfies $r(\theta,
\phi) \in \bk{0.7, 1.3}$ for all $(\theta, \phi)$.  

Our training dataset consists of $320$ manifold shapes having varying levels of
complexity.  We also hold out $40$ manifold shapes for testing.  For all
shapes, once a parameterization is chosen, we generate a mesh and a
quasi-uniform sampling consisting of $100,000$ points on the surface using the
PyACVD package from PyVista \cite{Sullivan2019,Valette2008}. The generated
meshes are used only for sampling purposes and visualizations.  For the
supervised training tasks for extracting geometric quantities,  we also compute
for these manifolds as a reference the unit normals, the first and second
fundamental forms, and Gaussian curvatures.

We choose the neighborhoods $\mathcal{N}_\epsilon(\bsy x)$ by using randomly
selected center points $\bsy x$ and querying all points within the
$\epsilon$-neighborhood.  As discussed above, we use a user specified number of
nearest neighbors $k$ which is computed in practice using a KD-tree. To ensure
accurate approximation of the geometry near the boundary of
$\mathcal{N}_\epsilon (\bsy x)$, we include all points within radius
$1.5\epsilon$ of $\bsy x$.  This is used as initial input to the model
$\mathcal{G}_\theta$, targeting predictions for output on all points
within the neighborhood $\set{\pn{u_i, v_i, w_i}}_{i=1}^N \in 
\mathcal{N}_\epsilon(\bsy x)$.

When making comparisons of accuracy between the GNP predictions
$\mathcal{G}_\theta$ and the ground truth geometric quantities, we use the
local coordinate system provided by the GNPs.  In particular, for the first two
fundamental forms $\set{\FirstForm}_{i=1}^N, \set{\SecondForm}_{i=1}^N$ we
convert the coordinate-dependent values based on spherical harmonics to the
corresponding coordinate-dependent values in the local frame given by
$\set{\bsy \psi_1, \bsy \psi_2, \bsy \psi_3}$.  For some of our studies, we
also use in the loss function the matrix inverse of the first fundamental form
$\FirstForm_i^{-1} = \{g^{ij}\}$, where the inverse of the metric tensor $g$ is
denoted by $g^{ij}$.  The inverse metric  plays an important in the
approximation of differential operators, such as the Laplace-Beltrami operator.  
This provides a set of geometric quantities for evaluating the accuracy of the
GNP predictions important in many geometric procedures and operators on
manifolds.  

When training the GNPs, we use the following loss function
\begin{equation}
    \label{equ_loss}
\begin{split}
\mathcal{L}(\mathcal{N}_\epsilon(\bsy x); \theta) & = 
\mathcal{L}_{\text{rel}}\pn{\set{\hat s(u_i, v_i)}, \set{w_i}}   
+ \lambda_1 \mathcal{L}_{\bsy\eta}\pn{\set{\bsy{\hat \eta}_i}, \set{\bsy\eta_i}}   
+ \lambda_2 \mathcal{L}_{\text{rel}}\pn{\set{\hat\FirstForm_i^{-1}}, \set{\FirstForm_i^{-1}}}  \\ 
&+ \lambda_3 \mathcal{L}_{\text{rel}}\pn{\set{\hat\SecondForm_i}, \set{\SecondForm_i}}  
+ \lambda_4 \mathcal{L}_{\text{rel}}\pn{\set{\hat K_i}, \set{K_i}}.
\end{split}
\end{equation}
The relative loss terms $\mathcal{L}_{\text{rel}}$ are defined as 
\begin{equation}
    \label{equ_relative_loss}
    \mathcal{L}_{\text{rel}}(f, g) = \frac{\norm{f - g}_2}{\norm{f}_2}.
\end{equation}
For the normals $\bsy{\eta}$, we use
\begin{equation}
\label{equ_normal_loss}
    \mathcal{L}_{\bsy\eta}\pn{\set{\bsy{\hat \eta}_i}, \set{\bsy\eta_i}}
    = \frac{1}{N} \sum_{i=1}^N \pn{1 - \bsy{\hat\eta_i} \cdot \bsy \eta_i }.
\end{equation}
The $\hat s(u, v) = \mathcal{G}_\theta(\mathcal{N}_\epsilon(\bsy x))$ denotes
the function constructed by the GNPs for parameters $\theta$.  The $\lambda_k$
provide parameters for adjusting the relative strength of the different
contributions to the loss.  Typical values we use in our training are
$\lambda_n = 0.5$ for $n = 1, 2, 3, 4$.  When training with a batch size
greater than $1$, we compute the mean of this loss over each of the
neighborhoods.   

In our training protocols, we consider both the case of no noise and two cases
with different types of noise.  This includes (i) uniform perturbations of
points obscuring the geometry, and (ii) large outlier points corrupting the
geometric sampling.  In the first case, we consider Gaussian noise that is
applied to every point with a fixed standard deviation.  In the second case, we
consider outliers that are generated by applying to $10\percent$ of the points a
large Gaussian noise.  This allows for GNPs to be trained to be robust to the
uniform noise and the outlier noise.   This is done by requiring that the
training be accurate for all neighborhood points $\mb{x}$ that are not
outliers.  This serves to signal the GNPs to learn to ignore the misleading
outlier points.
In the uniform noisy cases, we require the GNPs to give accurate results for
across all neighborhood points when evaluating the loss.  In this way, we can
introduce deliberately into the training dataset a set of artifacts that
obscure the underlying geometry and for which the GNPs need to compensate to
obtain robustness results.  Other types of noise and artifacts also can readily
be incorporated into our training protocols for the GNPs.  Our approach
provides ways to learn data-driven filtering and compensations for noise and
other artifacts in the point-clouds.

\section{Results}
\label{sec_results}

We learn GNPs using our training datasets based on radial manifold shapes from
our spherical harmonics methods discussed in Section~\ref{sec_learning_gnps}.
We show how our pre-trained GNPs are transferable to handle new geometries,
topologies, and tasks beyond the training datasets.  We demonstrate how the GNP
methods can be used for (i) estimating metrics and curvatures of surfaces, (ii)
deforming shapes driven by mean-curvature flows, and (iii) developing numerical
methods for solving geometric PDEs.

\subsection{Accuracy, Robustness, and Transferability of GNP Estimators 
for Geometric Quantities}
\label{sec_validation}

We show how the pre-trained GNPs can be used beyond the radial manifold shapes in 
the training datasets by considering a class of toroidal manifolds.  We consider
toroidal shapes diffeomorphic to a torus with parameterizations of the form
\begin{equation}
    \label{equ_toroidal manifold}
    \bsy \sigma(u, v) = 
    ((a(u, v)\cos(v) + b(u, v))\cos(u), 
    (a(u, v)\cos(v) + b(u, v))\sin(u),
    a(u, v)\sin(v)),
\end{equation}
where  $u, v \in [0, 2\pi)$.  The standard torus is obtained when $a, b$ are constant.
More generally, we obtain shapes by using functions $a(u, v), \ b(u, v)$ of the form
\begin{equation}
    \label{equ_torus radii}
    \begin{split}
        a(u, v) = a_0 + r_0 \sin(A_0 u)\cos(B_0 v),\;\; 
        b(u, v) = b_0 + r_1 \sin(A_1 u)\cos(B_1 v).
    \end{split}
\end{equation}
We sample shapes using random variates with $r_0 \sim \mathcal{U}(0.05a_0,
0.2a_0)$, $r_1 \sim \mathcal{U}(0, 0.08b_0)$, where $a_0=\tfrac{1}{3},
b_0=\tfrac{2}{3}$.  We also randomly select $A_i \in \set{1, 2, \dots, 5}$ and
$B_i \in \set{3, 4, \dots, 7}$.  The $\mathcal{U}[c_1,c_2]$ denotes the uniform
distribution on the interval $[c_1,c_2]$.  For testing the predictions of the
trained GNP models, we use that these shapes have explicit parameterizations in
a single coordinate chart.  This allows for comparisons between the pre-trained
GNPs and the geometric quantities when computed numerically using the
parameterization in equation~\ref{equ_toroidal manifold}.  We give the
expressions needed for numerical calculations of the geometric quantities in
Appendix~\ref{appendix_A}.

\begin{table}[h!]

\begin{tabular}{| l | c | c | c | c | c | c |}
\hline
\rowcolor{atz_table3}
\textbf{model} & \textbf{k} & \textbf{normal} & \textbf{metric inverse} &
\textbf{shape} & \textbf{gaussian}  & \textbf{mean} \\
\hline
\rowcolor{atz_table2} 
\multicolumn{7}{|l|}{\textbf{radial manifolds}} \\
\textbf{gnp, no-noise}  & 30 & 2.04e-06 & 6.57e-04 & 1.89e-02 & 4.07e-02 & 1.93e-02 \\
\rowcolor{atz_table1}
\textbf{gmls, no-noise}  & 30 & 3.11e-08 & 5.95e-05 & 4.10e-03 & 1.92e-02 & 7.15e-03 \\
\textbf{gnp, no-noise} & 50 & 4.55e-06 & 7.99e-04 & 1.70e-02 & 4.56e-02 & 2.24e-02 \\
\rowcolor{atz_table1}
\textbf{gmls, no-noise} & 50 & 1.89e-07 & 1.16e-04 & 7.06e-03 & 3.17e-02 & 1.21e-02 \\
\textbf{gnp, outliers,} $\sigma=$5e-03, & 50 & 3.70e-05 & 1.97e-03 & 2.75e-02 &
6.26e-02 & 3.14e-02 \\
\rowcolor{atz_table1}
\textbf{gmls, outliers,} $\sigma=$5e-03, & 50 & 3.39e-04 & 7.34e-03 & 3.72e-01
& 1.01e+00 & 4.11e-01 \\
\textbf{gnp, noise,} $\sigma=$1e-03, & 70 & 6.10e-05 & 5.62e-03 & 1.32e-01 &
1.75e-01 & 1.03e-01 \\
\rowcolor{atz_table1}
\textbf{gmls, noise,} $\sigma=$1e-03, & 70 & 6.11e-04 & 1.79e-02 & 1.17e+00 &
2.19e+00 & 7.00e-01 \\
\rowcolor{atz_table2} 
\multicolumn{7}{|l|}{\textbf{toroidal manifolds}} \\
\textbf{gnp, no-noise}  & 30 & 6.19e-07 & 3.79e-05 & 2.30e-02 & 4.03e-02 & 2.50e-02 \\
\rowcolor{atz_table1}
\textbf{gmls, no-noise}  & 30 & 3.97e-08 & 1.27e-06 & 5.90e-03 & 2.10e-02 & 8.34e-03 \\
\textbf{gnp, no-noise} & 50 & 1.43e-06 & 6.07e-05 & 2.48e-02 & 4.38e-02 & 3.34e-02 \\
\rowcolor{atz_table1}
\textbf{gmls, no-noise} & 50 & 3.97e-10 & 3.61e-03 & 1.03e-02 & 2.55e-02 & 1.49e-02 \\
\textbf{gnp, outliers,} $\sigma=$5e-03 & 50 & 4.16e-05 & 2.11e-04 & 3.71e-02 &
6.31e-02 & 4.45e-02 \\
\rowcolor{atz_table1}
\textbf{gmls, outliers,} $\sigma=$5e-03 & 50 & 2.43e-04 & 3.98e-04 & 3.58e-01 &
1.41e+00 & 3.94e-01 \\
\textbf{gnp, noise,} $\sigma=$1e-03 & 70 & 7.80e-05 & 6.25e-04 & 1.54e-01 &
2.16e-01 & 1.19e-01\\
\rowcolor{atz_table1}
\textbf{gmls, noise,} $\sigma=$1e-03 & 70 & 8.74e-05 & 1.25e-03 & 1.25e+00 &
2.24e+00 & 6.93e-01\\
\hline
\end{tabular}
\caption{\textbf{Accuracy and Robustness of GNPs to Noise and Outliers.}
We show results for the radial and toroidal manifolds for
predictions of our trained GNPs compared with GMLS estimators.
These are compared when varying the number of nearest neighbors $k$ 
for determining $\mathcal{N}_\epsilon(\bsy x)$ and in the presence 
of noise perturbations or outliers.  The Gaussian noise in 
these cases has standard deviation denoted by $\sigma$.  We show  
example point-cloud data with noise
in Figure~\ref{fig_noise}.}
\label{table_validation}
\end{table}

As a further study of the pre-trained GNP methods, we also make comparisons with 
alternative numerical methods based on Generalized
Moving Least Squares (GMLS)~\cite{Atzberger2020,mirzaei2012generalized}.  
We showed in prior work that GMLS can used for both 
geometric estimation and surface function approximation on scattered data 
by solving a collection of local least squares problems~\cite{Atzberger2020}.  In
these GMLS methods, at each point $\bsy x_i$ in the point cloud we
consider a neighborhood 
$\mathcal{N}_\epsilon (\bsy x_i) = \set{\bsy x_j}_{j=1}^n$ 
and solve the least-squares problem
\begin{equation}
    \label{equ_gmls_surface}
    q^*(u,v) = \argmin_{q \in \Phi} \sum_{j=1}^n
    \pn{s(u_j, v_j) - q(u_j, v_j)}^2 w\pn{\norm{(u_j, v_j)}_2}.
\end{equation}
Similar to the GNP methods, the GMLS estimator for geometric quantities 
at $\bsy{x_i}$ are computed by using for $q(u_j,v_j)$ 
the local coordinates $(u_j, v_j, s(u_j, v_j))$ as in
equation~\ref{equ_patch_eq} and from $q^*(u,v)$ using 
the expressions in Appendix~\ref{appendix_A}.
The approximation space $\Phi$ is taken to be   
polynomials up to degree $n=3$.  For the weight function $w$, 
we use $w(r) = (1 - r)_+^4$ where $(z)_+ = \max(z, 0)$.  For
more details on GMLS, see~\cite{Atzberger2020}.

While GNPs and GMLS use some similar geometric representations, there are a few key
differences between the approaches.  The GNP methods utilize a data-driven
neural operator approach to avoid needing to solve least-squares problems at
each evaluation site $\mb{x}_i$.  The GNP methods also allow for more general
loss functions which can through training target local and more 
global geometric features that
impact the latent geometric representations beyond
only targeting local least-squares reconstructions of shape or other
locally known functionals at the time of estimation.  In addition, the GNP methods 
can be trained to learn latent information for 
filtering out noise and outliers.  Data-driven filtering of
empirical artifacts and other idiosyncratic behaviors of particular types of
measurements or datasets can be readily incorporated into our
GNP training through empirical examples, data augmentations, 
and other protocols.  As we discuss in our examples, this 
provides unified strategies for utilizing GNPs to further impact the 
latent geometric representations to capture relevant information 
for different types of geometric tasks involving estimators 
and operators for point-clouds and manifolds.  

To validate the methods for important geometric quantities associated with
metric and curvature, we perform 
studies comparing the pre-trained GNPs, GMLS, and analytic results.  We consider
both radial and toroidal manifolds to show performance across different topologies.
These test cases also demonstrate how the methods perform on 
in-distribution shapes and out-of-distribution shapes relative to the 
training dataset based on radial manifolds.  We consider $40$ 
manifolds for each case of manifold.  We characterize the 
methods using the same relative loss terms as used in training when evaluating
the performance of the methods on the test data.  The loss terms 
are computed over neighborhoods of the manifolds for each point $\mb{x}_i$
and averaged across the $40$ manifolds.  In our comparisons, we compute  
the accuracy of the (i) shape reconstruction, (ii) normals, (iii) metric inverse, (iv) gaussian curvature, and 
(v) mean curvature.  We emphasize that the mean curvature $H$ was not included in 
the training loss, which provides an indication of the accuracy of the 
composite Weingarten map $\bsy W = \FirstForm^{-1}\SecondForm$, since 
$H = \mbox{trace}(\bsy{W})$.  We report our results for the methods
with and without noise in Table~\ref{table_validation}.

\begin{figure}[h!]
    \centering
    \includegraphics[width=0.8\linewidth]{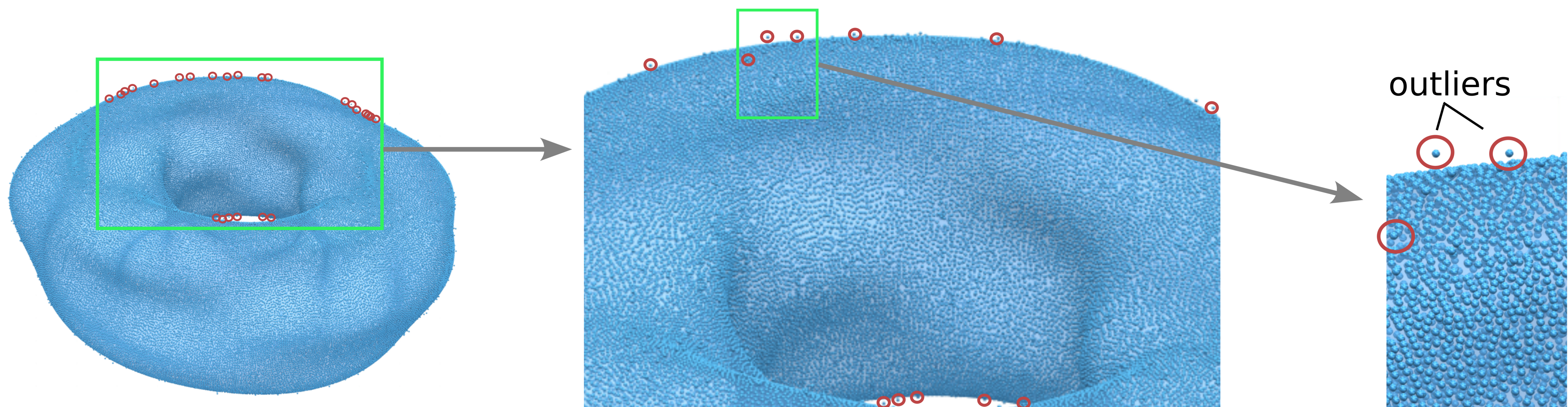}
    \caption{\textbf{Point-Cloud Data with Artifacts.}  The underlying 
geometry can be obscured by noise when working with point-cloud representations.  
We focus on the case of non-uniform samplings and 
outlier artifacts.  For an example shape sampled with
approximately $10\percent$ outliers \textit{(left)}, we 
highlight a subset of the outliers in the
data by circles \textit{(middle)}.  The shape also exhibits non-uniform sampling 
as can be seen in the most magnified view \textit{(right)}.  
These and other artifacts can be introduced into our datasets for training 
to enhance the robustness of the GNP methods.  }
    \label{fig_noise}
\end{figure}

We find for the no-noise and noisy cases that the GNPs are capable of learning
accurate estimators for the geometric quantities.  In the no-noise cases with
$k=30$ and $k=50$, we find while both methods give accurate results 
for both the radial and toroidal manifolds, the GMLS
estimators give more accurate results.  This is somewhat expected given that
GMLS solves a local least-squares objective at each surface location $\mb{x}_i$
specialized for targeting local shape reconstruction.
Both the GNPs and GMLS methods provide accurate estimators with errors less
than $5\percent$.  

In the case of uniform noise and outliers, we find the GNPs perform
significantly better than the GMLS methods.  In the outlier case, we use $k=50$
which provides additional samples for estimating the geometry.  We find that GNPs are
still able to reconstruct the radial shapes with an accuracy of $2.75 \percent$ while
GMLS only to an accuracy of $37\percent$.  Similarly, GNPs are able to obtain
accurate estimates despite outliers for gaussian curvature with error $6.25\percent$
and for mean curvature with error $3.14\percent$.  The GMLS estimates have  
gaussian curvatures with error $101\percent$ and mean curvatures with error $41.1\percent$,
see Table~\ref{table_validation}.
The GNPs also outperform GMLS for the toroidal manifolds
with outliers.  The GNPs have errors in reconstructing the shape $3.71\percent$,
gaussian curvature $6.31\percent$, mean curvature $4.45\percent$.  The GMLS methods have
errors in reconstructing the shape $35.8\percent$, gaussian curvature $140\percent$, mean
curvature $39.4\percent$. In the case of uniform Gaussian noise, we also find that
GNPs are more robust than the GMLS methods for both the radial and toroidal
manifolds.  The GNPs have errors respectively for the radial and toroidal
manifolds for the shape reconstruction $13.2\percent$, $15.4\percent$, gaussian curvature
$17.5\percent$, $21.6\percent$ and mean curvature $10.3\percent$, $11.9\percent$.  The GMLS methods
have errors for shape reconstruction $117\percent$, $125\percent$ and inaccurate estimates
for the gaussian and mean curvatures.

The results show the utility in GNPs of being able to use data-driven methods
to train the operators for mapping point-cloud samples $\{\mb{x}_i\}$ to latent
geometric features.  The GNP methods also can accommodate through training 
other types of noise and artifacts arising in practice for point-clouds 
without the need to design auxiliary
regularization terms.  In GNPs can be trained using empirical artifacts by incorporating 
them directly into the training datasets, performing data augmentations,
or using other protocols.  In this way, the GNPs can be used 
to learn robust estimators for geometric quantities of   
point-cloud representations obtained in practice.  The results also show the transferability 
of the GNP estimators which were trained on a dataset of radial manifolds
and were found to also provide accurate and robust results for the 
out-of-distribution toroidal shapes.  This allows for GNPs to be pretrained
and transferred for use in other data processing pipelines and  
computational methods.

\subsection{GNPs for Shape Deformations based on Mean-Curvature Flows} 
\label{sec_mean_flow}
As a further demonstration of how our pre-trained GNP methods can be transferred 
for use in other tasks, we perform computations of shape deformations based on 
Mean-Curvature Flows (MCF).  We show how the GNPs can be used as 
part of the numerical methods for these flows.  We consider deformations 
of an initial smooth manifold $\mathcal{M}$ that has an  
immersion given by the map $\varphi_0: \mathcal{M} \to \mathbb{R}^3$.  
The mean curvature flow (MCF) of $\mathcal{M}$ 
is a family of smooth immersions $\varphi_t : \mathcal{M} \to \R^3$
for $t \in [0, T)$, where $\varphi(p, t) = \varphi_t(p)$ is the
solution of the PDE 
\begin{equation}
    \label{equ_mean_flow}
    \begin{cases}
    \frac{\partial}{\partial t} \varphi(p, t) = H(p, t) \bsy \eta(p, t), 
    & p \in \mathcal{M}, t \in [0, T), \\
    \varphi(p, 0) = \varphi_0(p), & p \in \mathcal{M}.
    \end{cases}
\end{equation}
The mean curvature at $p \in \mathcal{M}$ is denoted by $H(p,t)$
and the outward unit normal by $\bsy \eta(p, t)$.  This 
flow deforms the surface by moving points $p$ in the 
direction of the normal vector $\bsy{\eta}$ at a 
rate proportional to the mean curvature $H$. 

We numerically approximate the PDE in equation~\ref{equ_mean_flow} by 
discretizing time into steps $t_k$ with $t_k = k\Delta{t} + t_0$ with
time-step $\Delta{t}$ 
and treat the manifold geometry through a finite point-cloud
sampling $\{\mb{x}_i\}_{i=1}^N$.  At each time, we 
deform the surface by moving each sample point $\bsy x_i^n$ 
in the normal direction in proportion to the local mean curvature using
\begin{equation}
\label{equ_mean_flow step}
    \bsy x_i^{n+1} = \bsy x_i^n + \Delta t H(\bsy x_i^n) \bsy \eta(\bsy x_i^n).
\end{equation}
This requires estimation from the point-cloud of the mean curvature 
$H(\bsy x_i^n)$ and normals $\bsy \eta(\bsy x_i^n)$.  For this purpose,
we use our pre-trained GNPs to obtain estimates of $H$ and $\bsy{\eta}$
for each point and time-step. 

To help ensure stability of the numerical methods, we apply a 
smoothing step on the values of $H$ to obtain
\begin{equation}
    \label{equ_curvature smoothing}
    \tilde H(\bsy x_i^n) = C_i^n \sum_{\bsy x_j^n \in B(\bsy x_i^n; 3r_0)}
    w(\norm{\bsy x_j^n - \bsy x_i^n}_2) H(\bsy x_j^n),
    \qquad w(r) = \exp(-r^2 / (2r_0^2)).
\end{equation}
The $C_i^n$ are chosen to normalize the sum so that the weights $C_i^n
w\pn{\norm{\bsy x_j^n - \bsy x_i^n}_2}$ sums to $1$. In practice, we set $r_0 =
0.00667$. As time evolves, areas of negative mean curvature will shrink and
point samplings will become more dense.  Similarly, regions also can become more 
rarefied.  To help maintain uniformity of the points and avoid round-off errors, 
we sub-sampled in areas of high density to eliminate points when they get 
too close together by only keeping one representative of the cluster. After each 
time step, we sub-sample as needed to ensure points are always at least distance 
$r_1$ from it's nearest neighbor, and we set $r_1 = 0.005$. 
In our examples, we did not need to do anything to mitigate rarefaction 
of the points, which could in principle be handled by interpolation and resampling
using our local GNP surface reconstructions.  Our GNP estimators have some build-in 
robustness to density variations since they are already handled 
to some degree by our criteria for selecting neighborhood patch sizes based on 
the $k$ nearest neighbors criteria.  In this way we are able to use the GNPs as part
of the numerical methods for deforming the surface by mean curvature flow. 

\begin{figure}[h!]
    \centering
    \includegraphics[width=0.8\linewidth]{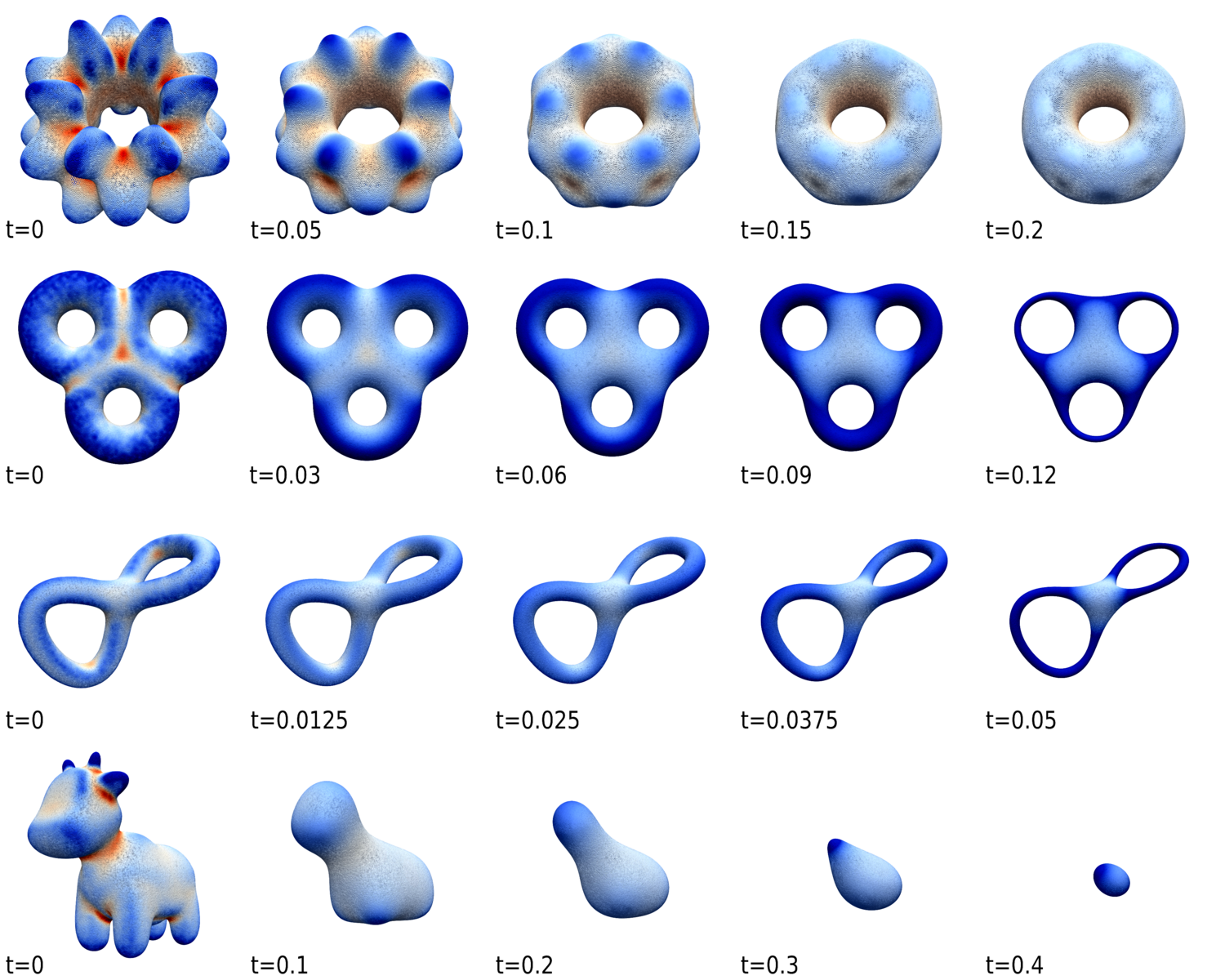}
    \caption{\textbf{Mean-Curvature Driven Flow.}
We show shape deformations evolved under mean-curvature flow (MCF) in
equation~\ref{equ_mean_flow} 
using our numerical methods based on pretrained GNPs.  
}
    \label{fig_mean_flow_shapes}
\end{figure}

To test the performance of the numerical methods for MCF 
based on the pretrained GNPs, we considered four different test shapes having
non-trivial geometries and topologies.  These consisted of shapes we refer to
as the (i) bumpy torus, (ii) fidget spinner (genus three shape), (iii) genus two
shape, and (iv) cow figurine.  The cow and genus shapes serve as common 
benchmark shapes used in the graphics community~\cite{Gao2017,Crane2013}.  
For MCF, it is known that the genus one and larger shapes can develop 
singularities.  For each of our shapes, we show results of the
mean-curvature flows up to the time of approach of a singularity
or when $N_*$ points get closer together than a threshold $\delta_*$. 
Another challenge for validating MCF is the lack of analytic solutions
for the shape deformations for non-trivial geometries.  We present
qualitative results for our GNP methods for MCF, and then present
quantitative results for numerical solvers for PDEs on manifolds
in the next section. 

We present our qualitative results for the MCF deformations for 
each of the shapes using
our GNP-based numerical methods, see Figure~\ref{fig_mean_flow_shapes}. 
In these results we used our pretrained 
GNPs with no-noise and $k=50$ nearest neighbors for estimating
$H, \bsy \eta$ at each $\bsy x_i^n$.  We use a time-step of 
$\Delta t = 10^{-4}$ and evolved each shape for at least $500$ 
iterations.  We find for each of the shapes the results are 
consistent with how we would expect mean-curvature flows to behave
and do not exhibit erratic estimates or deformations.
The GNP-based numerical methods consistently provide reliable time-steps
as indicated by the uniform and symmetric evolution of the bumpy torus
shape toward the expected target torus.  Similarly, the genus two
and three shapes show consistent evolution toward the singular shapes
expected from the mean curvature flows.  The cow figurine consists of
geometric features across several spatial scales and is found to evolve
toward the correct spherical target shape expected under mean-curvature 
flow.  The results show qualitatively how the GNPs can serve as reliable
components within other computational and numerical methods requiring   
geometric information.  To demonstrate more quantitative results, we 
consider next the development of numerical solvers for geometric PDEs
on manifolds.

\subsection{Using Transferable Pretrained GNPs for Developing Numerical Solvers
for Geometric PDEs} 
\label{sec_collocation}

We demonstrate the use of our pre-trained GNPs to develop numerical methods
for solving PDEs on manifolds.  We consider the Laplace-Beltrami (LB) equation
\begin{equation}
    \label{equ_laplace_beltrami_pde}
    \begin{cases}
        \Delta_{\text{LB}} \ u(\mb{x}) &= -f(\mb{x}), \\
        \int_{\mathcal{M}} u(\bsy x) \, d\bsy x &= 0,
    \end{cases}
\end{equation} 
where 
\begin{equation}
\label{equ_laplace_beltrami}
\Delta_{LB} = \frac{1}{\sqrt{\abs{g}}} \partial_i \pn{g^{ij} \sqrt{\abs{g}} \partial_j}.
\end{equation}
The $\Delta_{LB}$ generalizes the Laplacian to scalar functions on surfaces $\mathcal{M}$.
The $\partial_i$ denotes the derivative in the $i^{th}$ coordinate direction. 
The $|g|$ denotes the determinant of the metric tensor and $g^{ij}$
denotes the terms of the inverse metric.  The integral condition 
serves to provide a unique solution to the PDE for closed surfaces. 
A challenge that arises in  
development of effective numerical
methods is the ability to estimate from the manifold 
the geometric quantities $g$, $g^{ij}$, and derivatives.  For this purpose, we
develop collocation numerical
methods~\cite{fu2022localized,arnold1983asymptotic,moritz1978least} for
approximating solutions $u$ of the PDE that leverage geometric estimators
obtained from the pretrained GNPs.

In collocation numerical methods, approximate solutions $\tilde{u}$ to PDEs are
obtained by requiring the target differential relations $\mathcal{L} \tilde{u}(x_i)
= -f(x_i)$ hold approximately at a collection of evaluation points $x_i$.  The
function $\tilde{u}$ is represented by degrees of freedom $\alpha_k$ yielding
$\tilde{u} = \mathcal{U}[\bsy{\alpha}]$, where $\bsy{\alpha}$ is the vector
with $[\bsy{\alpha}]_k = \alpha_k$.  
Almost any form of interpolation or extension 
$\mathcal{U}[\bsy{\alpha}]$ for using the 
data to obtain a function $\tilde{u}$ can be utilized as long
as this provides increasingly accurate estimates of the action of the
differential operator $\mathcal{L} u$ as the density 
of collocation points 
$x_i$ increases~\cite{fu2022localized,moritz1978least,arnold1983asymptotic}.
Central to collocation methods is the ability to capture 
the local differential relations 
as the number of $\alpha_k$ and density
of points $x_i$ increases~\cite{arnold1983asymptotic}.  For this
purpose, we use for
$\mathcal{U}$ an approximate interpolation operation with $\alpha_k = u_k \approx
u(\tilde{x}_k)$, so $\tilde{u} = \mathcal{U}[\mb{u}]$ with $[\mb{u}]_k = u_k$.
The collection of points $\{\tilde{x}_k\}$ need not be the same as the collection
$\{x_i\}$. 

For the Laplace-Beltrami PDE, we let $\mathcal{L} = \Delta_{LB}$ and 
consider the differential relations 
\begin{equation}
    \label{equ_diff_relations1}
    \Delta_{\text{LB}} \tilde{u}(\mb{x}_i) = -f(\mb{x}_i) + r_i, \;\;\mbox{with}\;\;
r_i[\mb{u}] = \Delta_{\text{LB}} \tilde{u}(\mb{x}_i) - \left(-f(\mb{x}_i) \right).
\end{equation}
The residuals are denoted by $r_i$ which we will aim to make as small as possible. 
We use this to approximate the solution of the PDE in equation~\ref{equ_laplace_beltrami_pde} 
by $\tilde{u}^*(x) = \mathcal{U}[\mb{u}^*](x)$
where $[\mb{u}^*]_k = u_k^*$ is the solution of the following least-squares problem 
\begin{eqnarray}
\label{equ_ls_col}
\mb{u}^* 
= \arg\min_{\mb{u}} \sum_{i=1}^N \left(r_i[\mb{u}]\right)^2.
\end{eqnarray}

For the operator producing $\tilde{u} = \mathcal{U}[\mb{u}]$, we use GMLS to fit locally
polynomials by solving at each $\mb{x}_i$ the least-squares problem 
\begin{equation}
    \label{equ_gmls_p_star}
    p_i^*(\cdot) = \argmin_{p \in \mathbb{P}} \sum_{k=1}^n
    \pn{p(\bsy y_k) - u_k}^2 w\pn{\norm{\bsy y_k - \bsy x_i}_2},
\end{equation}
where $w$ is given by $w(r) = (1 - \tfrac{r}{\epsilon})_+^4$ and 
$\mathbb{P}$ is a space of Legendre 
Polynomials~\cite{boyd2001chebyshev,Trefethen2000} of degree $d=3$. 
This gives at each $\mb{x}_i$ a polynomial $p_i(\cdot)$ and we can evaluate the 
function as $\tilde{u}(\mb{x}_i) = p_i(\mb{x}_i)$.  

We also use the 
local polynomial representation to evaluate the action of differential 
operators $\mathcal{L}$.  We compute $\mathcal{L}v = \partial_jv$ for a 
function by letting 
$v_k = v(\mb{x}_k)$
and approximating 
$\partial_j v(\mb{x}_i)$ by $\partial_j p_i(\mb{x}_i)$,
where $p_i$ solves equation~\ref{equ_gmls_p_star} for $u_k = v_k$.
  We can 
compose derivative operations by repeating these approximations
successively using the output of the previous operations for the next sampled
input function $v$.  

For the Laplace-Beltrami $\mathcal{L} = \Delta_{LB}$ operator, 
further information is required beyond the derivatives $\partial_j$ 
since there are also contributions from the geometry, see
equation~\ref{equ_laplace_beltrami}.  For the geometric contributions to
the differential operator, we use our pretrained GNPs to obtain $g$, $g^{ij}$,
and other geometric terms.  We evaluate the action of the differential 
operator $\Delta_{LB} \tilde{u}(\mb{x}_i)$ by composing the evaluations
of $\partial_j$ with these geometric terms from the pretrained GNPs.  This 
provides for any choice of $u_k$ an approximation to the action of 
$\Delta_{LB}$ for computing the residuals in the collocation method in 
equation~\ref{equ_diff_relations1} and for solving the minimization 
problem in equation~\ref{equ_ls_col}.  

\begin{figure}[h!]
\label{fig_colloc}
\centering
\includegraphics[width=0.8\linewidth]{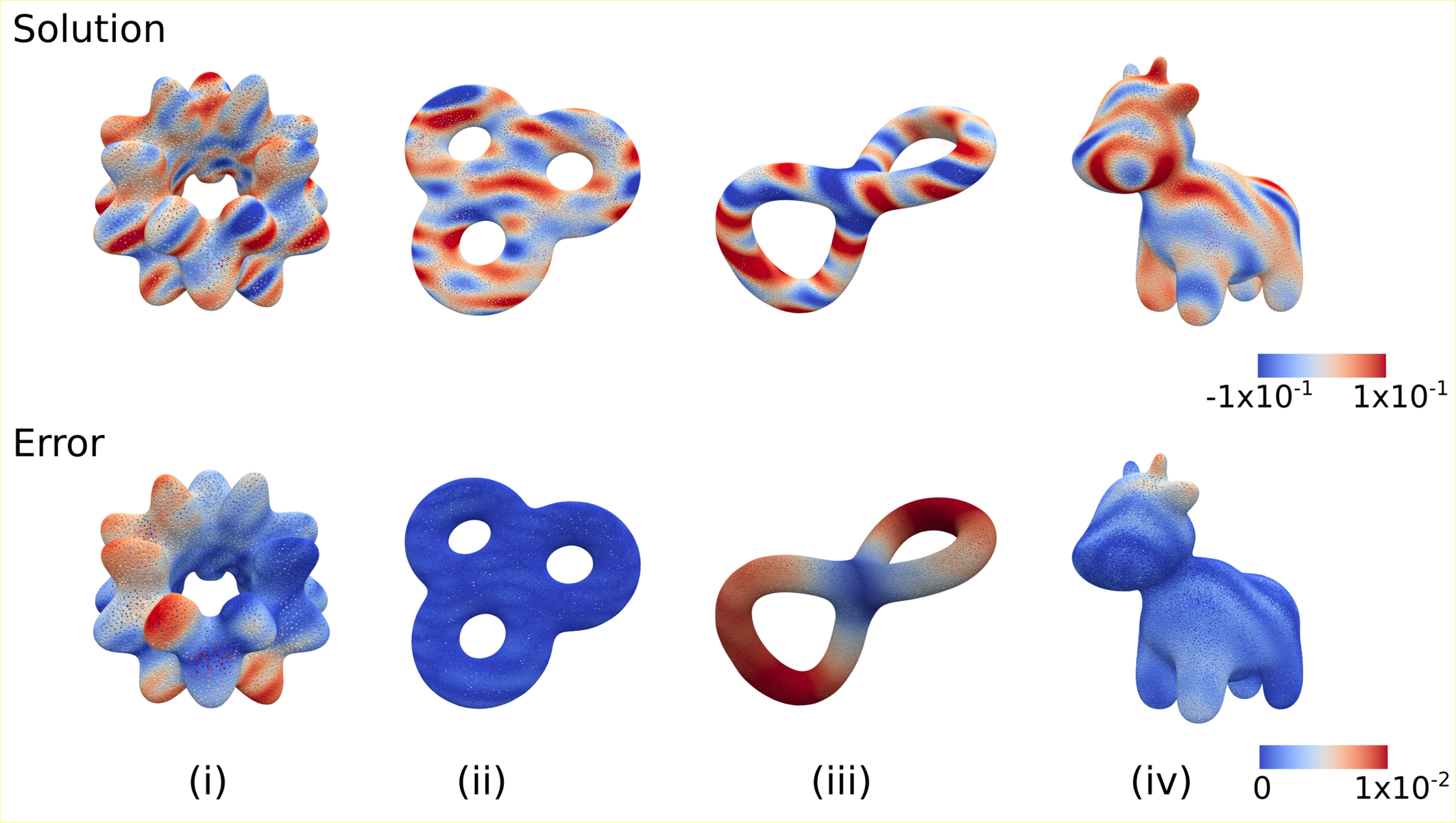}
\caption{\textbf{Geometric PDE Solvers based on GNPs.}  We show 
solutions of the Laplace-Beltrami PDE for
numerical methods based on our pre-trained GNPs. 
}
\end{figure}

To develop practical numerical methods for our collocation approach, 
we use 
that the operations for evaluation of $\Delta_{LB}$
are linear in $\mb{u}$.  This allows us to collect 
terms together to express the problem as seeking a 
solution to the linear system 
\begin{equation}
\label{equ_linear_sys}
A \mb{u} = \mb{f},
\end{equation}
where $[\mb{u}]_k = u_k \approx u(x_k)$, $[\mb{f}] = f_k \approx f(x_k)$,
and $[A\mb{u}]_k \approx \Delta_{LB} \tilde{u}(\mb{x}_k)$.
The stiffness matrix $A$ is obtained by composing the normal equations
for the GMLS least-squares problem in equation~\ref{equ_gmls_p_star} and the geometric 
contributions to $\Delta_{LB}$ from the pretrained GNPs in equation~\ref{equ_laplace_beltrami}.
Since the linear system in equation~\ref{equ_linear_sys} will typically be over-determined in 
our collocation methods, we solve for the $\mb{u}$ that
minimizes the residual $\|\mb{r}\|^2 = \|A \mb{u} - \mb{f}\|^2$.  
This is obtained by solving for $\mb{u}$ in the normal equations 
\begin{equation}
A^T A \mb{u} = A^T \mb{f}.
\end{equation}
To numerically approximate the solution of these equations, we develop
iterative methods based on Scipy's LGMRES solver~\cite{gommers2024scipy}.  
To improve the convergence rate of our 
iterative methods, we design algebraic multigrid preconditioners using
the package PyAMG \cite{pyamg2023}. 

For point-cloud methods, PDE solvers
for many applications need to be able to deal with the presence of noise
perturbations, outliers, or other artifacts.  We show how our GNP approaches
can be used to develop filters 
for PDE solvers to deal with noise.  
We found that the collocation
numerical methods perform quite poorly in the presence of outliers.  This is a
consequence of points moving not only in the normal direction, but also in the
tangential directions of the manifold resulting in solution distortions.  This
causes significant errors in approximating the function values on the manifold shape
especially at the outlier points. 

We introduce the following GNP filter method for identifying and processing outlier
points to improve the accuracy of the collocation methods.  We perform an
initial local surface reconstruction using our GNP methods on all points, 
since they have been trained to ignore outliers.  If we find for the local 
GNP predicted shape reconstructions that a point $\mb{x}_i$ with coordinate 
$(u_i,v_i,w_i)$ has a deviation value $\abs{w_i - s(u_i, v_i)} > \alpha$
above the threshold $\alpha$, we remove this point from the neighborhood during
further processing.
We use $\alpha = 0.1$ in practice.  We then use our methods to approximate $u$
and the remaining geometric quantities using the remaining points after 
the GNP filtering.

For our numerical methods, we use in our studies point-clouds with $100,000$
sample points for each shape.  When imposing the differential relations in
equation~\ref{equ_diff_relations1}, we evaluate the operator
$\Delta_{LB}\tilde{u}(\mb{x}_i)$ and the right-hand-side (RHS) $f(\mb{x}_i)$
at all $n=100,000$ sample points $\{\mb{x}_i\}_{i=1}^n$.  To ensure uniqueness
and robustness of the residual least-squares problem, we use a subset  $N =
0.9n = 90,000$ of the sample points for determining $\tilde u$ with the degrees
of freedom $u_k$ associated with each of the points in $\{\mb{x}_k\}_{i=1}^N$.
We use RHS functions $f(x)$ for testing the solvers by generating 
Fourier Series for $x \in \mathbb{R}^3$ with $M=10$ modes in each direction with 
coefficients drawn from a normal distribution and spatial period $[-1,1]$.  We 
then restrict these functions to the manifold shape.  We also normalize $f(x)$
so it has values in the range $[-20,20]$.  In each case,  
we test the solver in equation~\ref{equ_laplace_beltrami_pde} with 
$10$ samples of these RHS terms $f(x)$.

We study how our numerical methods based on the pretrained GNPs perform for
Laplace-Beltrami PDEs on geometries with and without noise.  We consider the
four shapes discussed in Section~\ref{sec_mean_flow}.  We refer to these shapes
as the (i) bumpy torus, (ii) genus three shape, (iii) genus two shape, and (iv)
cow figurine, see Figure~\ref{fig_mean_flow_shapes}.  We compare our results
for the GNP numerical methods with those that use as an alternative the GMLS
estimators for the geometric contributions, as discussed in
Section~\ref{sec_validation}.  When performing the studies with noise, we
employ our GNP filtering methods discussed above for the collocation methods in
both the GMLS and GNP cases.  If we did not perform filtering of outliers for
the collocation PDE solver, GMLS has large errors and does not produce viable 
outputs.  We show the results of our studies in Table~\ref{table_collocation}. 

\begin{table}[h!]
    \centering
    \begin{tabular}{| l | c | c | c | c | c |}
    \hline
    \rowcolor{atz_table3}
     \textbf{model} & \textbf{k} & \textbf{bumpy torus} & \textbf{genus-3} &
\textbf{genus-2} & \textbf{cow} \\
    \hline
    \textbf{gnp, no-noise}  & 30 & 7.97e-02 & 9.55e-03 &  8.23e-02 & 7.11e-02 \\
    \rowcolor{atz_table1}
    \textbf{gmls, no-noise}  & 30 & 8.03e-02 & 9.88e-03 &  7.33e-02 & 7.27e-02 \\
    \textbf{gnp, no-noise} & 50 & 8.49e-02 & 1.26e-02 & 3.86-02 & 9.93-02 \\
    \rowcolor{atz_table1}
    \textbf{gmls, no-noise} & 50 & 1.01e-01 & 1.39e-02 & 2.94-02 & 6.63-02 \\
    \textbf{gnp, outliers,} $\sigma=$5e-03 & 50 & 9.30e-02 & 1.72e-02 & 6.94e-02 & 1.08e-01  \\
    \rowcolor{atz_table1}
    \textbf{gmls, outliers,} $\sigma=$5e-03 & 50 & 9.12e-02 & 1.74e-02 & 4.98e-02 & 9.20e-02  \\
    \textbf{gnp, noise,}  $\sigma=$1e-03 & 70 & 9.95e-02 & 2.02e-02 & 5.52e-02 & 1.02e-01  \\
    \rowcolor{atz_table1}
    \textbf{gmls, noise,}  $\sigma=$1e-03 & 70 & 1.59e-01 & 6.09e-02 & 1.22e-01 & 9.99e-01  \\
    \hline
    \end{tabular}
    \caption{\textbf{Geometric PDE Solvers based on GNPs.}  We show results for
the accuracy and robustness of our PDE solvers using numerical methods based on
pretrained GNPs.  We also make comparisons with alternative GMLS methods.  
In both cases, we used our GNP methods for filtering outliers, otherwise the 
GMLS collocation methods do not produce viable results. 
We show cases when varying the neighborhood 
size $k$ and cases with and without noise. } 
    \label{table_collocation}
\end{table}

We find in the no-noise case that the collocation methods based on pre-trained
GNPs and GMLS methods behave comparably for $k=30$ and $k=50$.  While both
methods are accurate with a precision of at least $10\percent$ or better, we find for
$k=50$ the results are slightly less accurate.  This arises since as patch
sizes become larger they can smooth locally the geometric features of the
shapes.  In the outlier case, while the GMLS and GNP methods performed
similarly, we emphasize the GNPs were used as part of the pre-processing filter
for the later GMLS steps to help mitigate noise and outliers.  As mentioned,
if this was not done GMLS has errors that are too large to produce viable 
results in the collocation methods.  We see our GNP filtering
yields enhanced precision of the geometric and PDE estimates allowing
for similar accuracies for both solvers. 

In the case of uniform Gaussian noise, we find the GNPs performed consistently
better than the GMLS methods.  We see an especially significant difference for
the genus-2 shape.  The pre-trained GNP methods yield a solver with $5.5\percent$
accuracy compared to $12.2\percent$ for the purely GMLS-based methods.  We see similar
improvements for the genus-3 shape, where the GNP methods yield a solver 
with accuracy of $2.0\percent$ compared to $6.0\percent$ for the purely GMLS-based 
methods.  A notable aspect of each of these cases 
are regions having large curvatures requiring robust estimators for the 
geometric and PDE contributions.  This 
shows a key advantage of the learned GNP estimators. 
Since the GNPs were trained with artifacts during training, 
they are able to learn more robust estimators that can 
compensate for noise in the point-cloud data.  
The results of these studies show how the pre-trained GNPs can be
used in the development of 
numerical methods for approximating solutions of geometric 
PDEs.

\section*{Conclusions}
We have shown how transferable GNP models can be learned
for processing point-cloud representations for use in geometric tasks.  
The GNPs can be used to learn 
estimators for key geometric quantities and other features.  The GNP
methods also allow for training that incorporates data-driven filtering for
handling noise, outliers, and other artifacts in non-pristine point-clouds.  
We demonstrated how the 
pretrained GNPs can be used in tasks that include (i) robustly estimating
geometric quantities related to the metric and curvatures, (ii) 
tracking shape changes driven by mean-curvature
flows, and (iii) developing numerical methods for approximating the solutions
of geometric PDEs.  The GNP models also can be incorporated
readily into other data processing pipelines and computational methods.  
We release an open source package with training codes and with weights 
for our pre-trained GNPs.  The introduced approaches provide methods for 
obtaining general transferable GNP models for performing geometric tasks.

\section*{Open Source Package}
We release an open source package for our methods at  \\
\url{https://github.com/atzberg/geo_neural_op}.

\section*{Acknowledgments}

Authors research supported by grant NSF Grant DMS-2306101.  Authors also would
like to acknowledge computational resources and administrative support at
the UCSB Center for Scientific Computing (CSC) with grants
NSF-CNS-1725797, MRSEC: NSF-DMR-2308708, Pod-GPUs: OAC-1925717, and support from
the California NanoSystems Institute (CNSI) at UCSB.  P.J.A. also would
like to acknowledge a hardware grant from Nvidia.

\newpage
\clearpage

\appendix
\addcontentsline{toc}{section}{Appendices}

\printbibliography

\newpage

\appendix

\section{Fundamental Forms, Curvature, and Operators on Surfaces}
\label{appendix_A}
We give expressions for how a few key geometric quantities can 
be computed from local parameterizations $\bsy \sigma(u, v)$ of 
the surface.  In the coordinates $(u, v)$, the 
first fundamental form $\FirstForm$ is defined as
\begin{equation}
\label{equ_first form}
\FirstForm = 
\begin{bmatrix}
\bsy \sigma_u \cdot \bsy \sigma_u & \bsy \sigma_u \cdot \bsy \sigma_v \\
\bsy \sigma_v \cdot \bsy \sigma_u & \bsy \sigma_v \cdot \bsy \sigma_v
\end{bmatrix}
 = \begin{bmatrix}
    E & F \\
    F & G
\end{bmatrix}.
\end{equation}
We denote the derivatives by $\bsy{\sigma}_u = \partial_u \bsy{\sigma} / \partial u$,
$\bsy{\sigma}_v = \partial_v \bsy{\sigma} / \partial v$, and similarly for higher-order
terms. 
This first fundamental form $\FirstForm$ is also equivalent to 
the metric tensor $\mb{g} = \FirstForm$.  The $\FirstForm$ can
be used for 
computations involving distances, arc lengths, and angles on the surface. 
The second fundamental 
form $\SecondForm$ is defined as
\begin{equation}
\label{equ_second form}
\SecondForm =
\begin{bmatrix}
    \bsy \sigma_{uu} \cdot \bsy n & \bsy \sigma_{uv} \cdot \bsy n \\
    \bsy \sigma_{uv} \cdot \bsy n & \bsy \sigma_{vv} \cdot \bsy n
\end{bmatrix}
 = \begin{bmatrix}
    L & M \\
    M & N
\end{bmatrix}.
\end{equation}
The $\bsy n$ denotes the outward unit normal to the surface given by
\begin{equation}
\label{equ_normal}
\bsy n = \frac{\bsy \sigma_u \times \bsy \sigma_v}{\norm{\bsy \sigma_u \times \bsy \sigma_v}}.
\end{equation}
These can be combined to obtain the 
Weingarten map $\bsy W = \FirstForm^{-1} \SecondForm$.  The $\bsy W$ can be used 
to compute the Gaussian curvature $K$ and the mean curvature $H$ of the surface. 
These are given by
\begin{equation}
\label{equ_gaussian curvature}
K = \det(\bsy W) = \frac{LN - M^2}{EG - F^2},
\end{equation}
\begin{equation}
\label{equ_mean curvature}
H = \frac{1}{2} \text{tr}(\bsy W) = \frac{1}{2} \pn{\frac{LG - 2MF + NE}{EG - F^2}}.
\end{equation}
The Laplace-Beltrami operator $\Delta_{LB}$ extends the Laplacian to
scalar functions on the surface.  The can be expressed as
\begin{equation}
\label{equ_LB2}
\Delta_{LB} = \frac{1}{\sqrt{\abs{g}}} \partial_i \pn{g^{ij} \sqrt{\abs{g}} \partial_j}.
\end{equation}
The $|g|$ denotes the determinant of the metric tensor and $g^{ij}$
denotes the terms of the inverse metric.  In these expressions we use
the Einstein tensor notations for implicit
summation~\cite{abraham2012manifolds}.  Further discussions of these
geometric quantities and computations also can be found
in~\cite{abraham2012manifolds,pressley2010elementary,Atzberger2018}.

\end{document}